\documentclass[11pt]{article}

\usepackage[final]{acl}

\usepackage{times}
\usepackage{latexsym}

\usepackage[T1]{fontenc}

\usepackage[utf8]{inputenc}

\usepackage{microtype}

\usepackage{inconsolata}

\usepackage{graphicx}
\usepackage{multirow}
\usepackage{booktabs}
\usepackage{graphicx}
\usepackage{float}
\usepackage{algorithm}
\usepackage{algpseudocode}
\usepackage{caption}
\usepackage{subfigure}
\usepackage{amssymb}
\usepackage{amsmath}

\usepackage{listings}
\usepackage{xcolor}

\newcommand\qacc{Strict Accuracy}
\newcommand\sacc{Valid Rate}
\newcommand\model{QualEQS}

\title{Quality Over Clicks: Iterative Reinforcement Learning for Early-Stage E-Commerce Query Suggestion}

\author{Qi Sun$^{1}$\textsuperscript{*}, Kejun Xiao$^{1}$ \thanks{Both authors contributed equally to this work.}, Huaipeng Zhao$^{1}$,Tao Luo $^{1}$,  
    Xiaoyi Zeng $^{1}$  \\
  $^{1}$Alibaba International Digital Commercial Group \\ 
  \texttt{\{qiran.sq,xiaokejunkejun.xia\}@alibaba-inc.com}, \\
  }

\begin{document}
\maketitle
\begin{abstract}
Existing dialogue systems rely on query suggestion to enhance user engagement. Recent approaches mainly optimize generative models using click-through rate (CTR) models to align with user preferences. However, these methods are less effective in early-stage deployment scenarios, where click feedback is sparse and insufficient for training a reliable CTR model.
To bridge this gap, we propose \textbf{\model}, a quality-first iterative reinforcement learning framework for e-commerce query suggestion. We formalize actionable suggestion quality along three dimensions that directly affect downstream usability: answerability, factuality, and information gain. To continuously improve from online traffic without click supervision, we further propose group-level disagreement among candidate suggestions to identify ambiguous query contexts and mine hard training cases for iterative refinement.
We also introduce \textbf{EQS-Benchmark}\footnote{The benchmark and code will be made publicly available after internal review.}, a dataset of 16,949 real-world e-commerce queries for offline training and evaluation. Experiments show that our quality-based offline metrics correlate strongly with online performance, providing a practical evaluation recipe for sparse-feedback deployment. In both offline and online settings, \model{} consistently outperforms strong baselines, yielding a 6.81\% improvement in online ChatPV in a real-world enterprise-level conversational shopping assistant system.

\end{abstract}

\section{Introduction}
Query suggestion (QS) is a key component of modern conversational assistants, especially in e-commerce, where users often arrive with vague, incomplete, or evolving intents. During multi-turn interactions, the assistant can proactively present clickable follow-up queries that help users refine their needs with minimal effort. 

Existing QS methods are mainly retrieval-based or generation-based~\cite{retrieval2018entity, retrieval2008context, Min2025ctr, bi2025consistency}. Retrieval-based approaches are constrained by historical query pools and often struggle to adapt to rapidly evolving user intents ~\cite{retrieval2023exploiting}. Generation-based approaches have used large language models (LLMs) to produce more flexible and context-aware suggestions, often optimized with click-based objectives such as click-through rate (CTR) rewards or preference signals derived from user interactions~\cite{bacciu2024generating, shen2024enhancing, wang2023zero, Min2025ctr}.

\begin{figure}
	\centering
	\includegraphics[width=0.25\textheight]{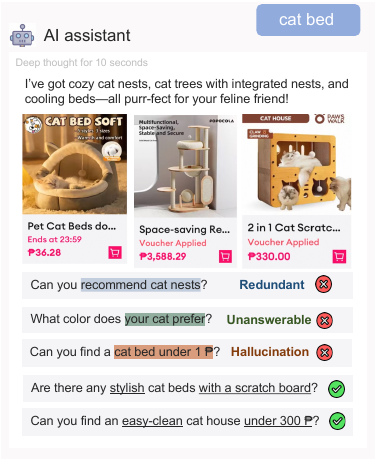}
    
	\caption{Example of an AI shopping assistant where query suggestions are presented as clickable elements.}
    \vspace{-0.9em}
	\label{fig:intro}  
 
\end{figure}

However, these methods are poorly suited to early-stage deployment. When a conversational shopping assistant is first launched, traffic is limited and click feedback is sparse. It is difficult to train a reliable CTR model or construct stable preference signals. For example, \citet{Min2025ctr} rely on more than 20 million interaction records, a scale that is rarely available in the early phase of product deployment. More fundamentally, clicks do not fully capture whether a suggested query is actually useful for downstream conversational agents. As shown in Figure~\ref{fig:intro}, a suggestion may attract clicks while still being unanswerable, factually misleading, or insufficiently informative, ultimately harming multi-turn interaction quality.

\begin{figure*}
	\centering
	\includegraphics[width=0.6\textheight]{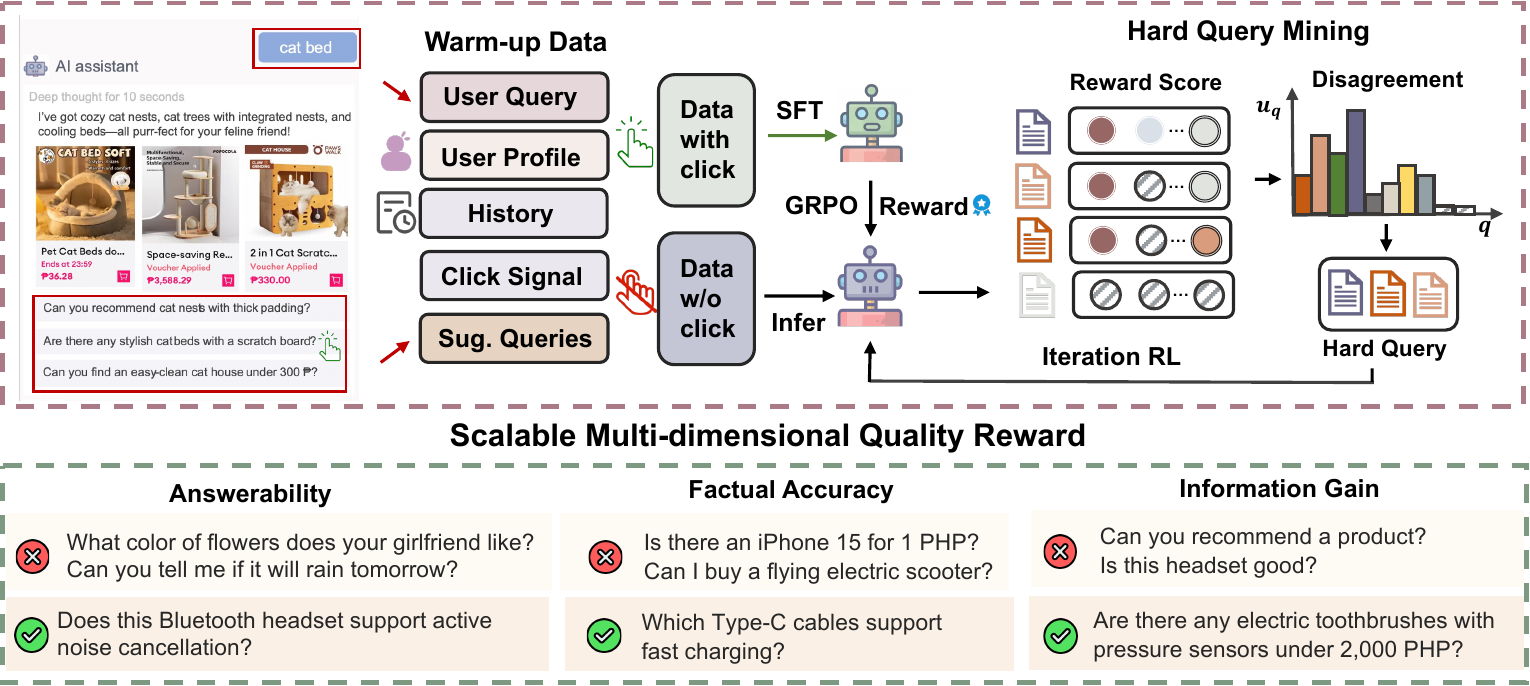}
	\caption{Overview of our \model{} framework: (1) warm-starting the policy on early clicked interactions, (2) optimizing it with quality-first rewards, (3) mining hard unlabeled online queries via group disagreement, and (4) iteratively refining the policy. }
	\label{fig:pipeline}  
  \vspace{-0.9em}
\end{figure*}

To bridge this gap, we propose \textbf{\model}, a quality-first iterative reinforcement learning (RL) framework for early-stage e-commerce query suggestion. Instead of relying on dense click supervision, \model{} optimizes actionable suggestion quality using three reward dimensions: answerability, factuality, and information gain. To exploit online traffic without click signals, we leverage group-level disagreement among candidate suggestions generated for the same query context. When the model produces suggestions with highly inconsistent quality for the same input, the underlying query context is more likely to be ambiguous, difficult, or weakly modeled. We therefore use this disagreement signal to mine hard online cases and iteratively improve the model.

We also ask a practical deployment question: \textit{what offline signal can reliably predict online gains before sufficient click feedback exists?} To support this setting, we introduce \textbf{EQS-Benchmark}, a dataset of 16,949 real-world e-commerce queries for offline training and evaluation. Experiments show that our quality-based offline metrics correlate strongly with online engagement, suggesting that intrinsic quality can serve as an effective surrogate objective in sparse-feedback deployment. Across both offline and online experiments, \model{} consistently outperforms strong baselines and improves online ChatPV by 6.81\%.

The main contributions of our work are summarized as follows:
\begin{itemize}
    
    \item We propose \textbf{\model}, an iterative reinforcement learning framework that operationalizes actionable suggestion quality using three task-specific dimensions: answerability, factuality, and information gain.
    
    \item We introduce a group-disagreement-driven mining strategy that leverages reward disagreement among candidate suggestions to identify ambiguous online queries and iteratively improve the model.
    
    \item We construct \textbf{EQS-Benchmark}, a real-world dataset for offline training and evaluation, and show that quality-based offline metrics correlate strongly with online engagement and business outcomes.
    
\end{itemize}

\section{Problem Formulation}
 We formulate query suggestion as conditional generation. Given an input $q$ consisting of the current user query, dialogue history, and user profile, the policy $M_\theta$ generates an ordered list of $k$ suggestions $(s_1, \dots, s_k)$. In our setting, offline training optimizes a composite quality reward:
\begin{equation}
\max_{\theta} \mathbb{E}_{q \sim \mathcal{Q}} \left[ \mathbb{E}_{( s_1,...,s_k ) \sim M_\theta(\cdot \mid q)} \left[ R( s_1,...,s_k) \right] \right],
\end{equation}
where $M_\theta$ is the generative QS model and $R$ is the quality reward.

\section{Methodology}
Figure~\ref{fig:pipeline} shows our framework, which consists of four stages: (1) warm-starting the policy on early clicked interactions, (2) optimizing the policy with quality-first rewards, (3) mining hard unlabeled online queries with group-disagreement-driven sampling, and (4) iteratively refining the policy.

\subsection{Warm Start SFT}
Although sparse early interactions are insufficient to train a reliable click-based reward model, the small subset of suggestions preferred by users still provides useful positive supervision for initializing the policy. We collect click-positive tuples $\mathcal{D}_{\text{click}} = \{(q, s_1, \ldots, s_k)\}$ from online traffic, where at least one suggested query is clicked by the user. We then fine-tune Qwen3-4B on $\mathcal{D}_{\text{click}}$:
\begin{equation}
M_{\text{SFT}} = \operatorname{Train}(M_{\text{init}},\, \mathcal{D}_{\text{click}}),
\end{equation}
where $M_{\text{init}}$ is Qwen3-4B~\cite{yang2025qwen3technicalreport}.

\subsection{Quality-First RL}
Directly optimizing for clicks may favor click-attractive but unhelpful suggestions. To avoid this mismatch, we optimize the policy with actionable quality reward. Starting from the warm-started policy $M_{\text{SFT}}$, we further train the model with GRPO~\cite{grpo}:
\begin{equation}
M_{\text{RL}} = \arg\max_{\theta}~\mathbb{E}_{(s_1,...,s_k)\sim M_\theta(\cdot\mid q)}[R(s_1,...,s_k)],
\end{equation}
where $M_\theta$ denotes the query suggestion policy and $R(\cdot)$ is the actionable quality reward defined over the generated suggestion set.

\subsection{Scalable Reward Modeling}
As shown in Figure \ref{fig:pipeline}, we operationalize actionable suggestion quality using three task-specific dimensions.

\paragraph{Answerability.} This measures whether a suggested query is within the scope of the platform and can be handled by the downstream assistant. It penalizes out-of-domain, privacy-sensitive, or assistant-perspective suggestions that a real user would not issue within the shopping scenario.
\paragraph{Factuality.} This dimension measures whether a suggested query is grounded in realistic product and commerce knowledge. It penalizes clearly false, misleading, or impossible content, such as unrealistic product prices or non-existent functionalities.
\paragraph{Information Gain.} This dimension measures whether a suggested query moves the interaction forward by refining or extending the current intent. It rewards suggestions that add useful constraints or context, such as product attributes, brands, usage scenarios, or specific budgets, rather than merely repeating the original query.

During training, we generate a set of $k$ candidate suggestions $(s_1, s_2, \dots, s_k)$ for each input context. Each candidate receives a quality score:
\begin{equation}
r(s_i) = r_{\mathrm{ans}}(s_i)\cdot r_{\mathrm{fact}}(s_i)\cdot r_{\mathrm{info}}(s_i),
\end{equation}
where $r_{\mathrm{ans}}(s_i)$, $r_{\mathrm{fact}}(s_i)$, and $r_{\mathrm{info}}(s_i)$ denote the answerability, factuality, and information gain rewards for the $i$-th suggestion, respectively. We further apply a format reward $r_f$ to enforce valid structured output, and define the set-level reward as:
\begin{equation}
r_q = r_f \cdot \frac{1}{k} \sum_{i=1}^{k} r(s_i).
\end{equation}

We use Qwen-30B-A3B as the reward evaluator, with few-shot examples constructed from annotated positive samples and real online bad cases collected during deployment. Compared with fixed heuristics, this design provides finer-grained judgments of answerability, factuality, and information gain, especially for subtle or long-tail failure patterns that are difficult to capture with smaller models in sparse-feedback settings. It is also easy to extend in practice: as new failure cases emerge from online traffic, they can be added to the in-context examples to continuously improve the evaluator without training a separate reward model. A sanitized version of the prompt is provided in Appendix~\ref{sec:appendix_C}.

\subsection{Hard Query Mining}
To exploit online traffic without click supervision, we mine hard query contexts from unclicked online data. For each online query context $q$, the current policy $M_{\text{RL}}$ generates a set of candidate suggestions $(s_1, \dots, s_k)$, each associated with an actionable quality score $r(s_i)$. We observe that when the model produces candidates with highly inconsistent quality for the same context, the input is often ambiguous, difficult, or weakly modeled. We therefore use group-level reward disagreement as a signal for hard-query mining.

Formally, we define the disagreement score of context $q$ as the variance of candidate quality scores:
\begin{equation}
u_q = \frac{1}{k} \sum_{i=1}^{k} \left( r(s_i) - \frac{1}{k} \sum_{j=1}^{k} r(s_j) \right)^2.
\end{equation}

Contexts with high disagreement scores are selected into a hard subset for the next round of RL training. 

\subsection{Iterative Quality-First Optimization}
As shown in Algorithm \ref{alg:gqs}, our training pipeline contains following stages: (1) \textit{warm start}, where we initialize the policy with supervised fine-tuning on clicked data $\mathcal{D}_{\text{click}}$; (2) \textit{quality-first RL}, where we optimize the policy with actionable quality reward; (3) \textit{hard-query mining}, where we apply the model to unclicked data $\mathcal{D}_{\text{unclick}}$ and select high-disagreement contexts as $\mathcal{D}_{\text{hard}}$; and (4) \textit{iterative training}, where we continue RL on both $\mathcal{D}_{\text{click}}$ and $\mathcal{D}_{\text{hard}}$ for $K$ rounds. This pipeline enables continual improvement under sparse feedback.

\begin{algorithm}
\caption{Iterative Training for \model{}}
\label{alg:gqs}
\textbf{Input:} Clicked interactions $\mathcal{D}_{\text{click}}$, unclicked online inputs $\mathcal{D}_{\text{unclick}}$, base model $M$, actionable quality reward $R$, iteration number $K$.\\
\textbf{Output:} Final query suggestion model $M_{\text{final}}$.

\begin{algorithmic}
\Require $\mathcal{D}_{\text{click}}, \mathcal{D}_{\text{unclick}}, K, M, R$
\Ensure $K > 0$

\State \textbf{1. Warm-start}
\State $M_{\text{SFT}} \gets Train(M, \mathcal{D}_{\text{click}})$

\State \textbf{2. Quality-first RL}
\State $M_{\text{RL}} \gets TrainRL(M_{\text{SFT}}, \mathcal{D}_{\text{click}}, R)$

\For{$t=1$ to $K$}
    \State \textbf{3. Hard query mining}
    \State $\mathcal{D}_{\text{cand}} \gets Predict(M_{\text{RL}}, \mathcal{D}_{\text{unclick}})$
    \State $\mathcal{D}_{\text{hard}} \gets Sampling(\mathcal{D}_{\text{cand}}, R)$

    \State \textbf{4. Iterative refinement}
    \State $M_{\text{RL}} \gets TrainRL(M_{\text{RL}}, \mathcal{D}_{\text{hard}}, \mathcal{D}_{\text{click}}, R)$
\EndFor

\State $M_{\text{final}} \gets M_{\text{RL}}$
\State \Return $M_{\text{final}}$
\end{algorithmic}
\end{algorithm}

\section{Experiments}
\subsection{Experimental Setup}
\paragraph{Datasets} We propose EQS-Benchmark for offline experiments. The training dataset consists of 5,914 click-positive and 10,535 click-negative samples. The test set contains 500 high-quality user queries, carefully selected to cover all major e-commerce scenarios and intents, including platform QA, product recommendations, and product QA. Details can be seen in Appendix \ref{sec:benchmark}.

\paragraph{Metrics}
 We employ Qwen3-Max to evaluate $k$ suggested queries for each user query under answerability, factual accuracy, and information Gain. We report \textbf{Strict Accuracy}, which is the fraction of user queries for which all $k$ generated suggestions satisfy all three quality criteria, and \textbf{Valid Rate}, which is the fraction of all generated suggestions that satisfy the criteria. We randomly sampled 500 generated query suggestions and compared human annotations with LLM-as-judge evaluations, achieving 90.40\% consistency. The online metric can be seen in Appendix \ref{sec:online_mtric}.

\paragraph{Online Deployment}
For production deployment, the final \model{} model is deployed on two L20 GPUs with an average online inference latency of 686\,ms. During RL training, we deploy Qwen-30B-A3B as the reward model on two H20 GPUs (141\,GB total GPU memory) for online reward scoring. This overhead is incurred only during training stage. More details are given in Appendix~\ref{sec:reward_deployment}.

\paragraph{Baselines}
Considering the stringent response time (RT) requirements in our scenario, we primarily evaluated models with fewer than 8B parameters, as well as flash/mini versions of both open-source and closed-source models including Qwen, Gemini, and OpenAI series. The training details can be seen in Appendix \ref{sec:appendix_D}.

\begin{table*}
\centering \small
\begin{tabular}{lccccc}
\toprule
 \multirow{2}*{\textbf{Models}} & \multirow{2}*{\textbf{\qacc}} & \multirow{2}*{\textbf{\sacc}} & \multicolumn{3}{c}{\textbf{Fine-grained Quality Accuracy}} \\
\cmidrule(lr){4-6}
 &  &  & \textbf{Answerable} & \textbf{Factuality} & \textbf{Informativeness} \\
\midrule
\multicolumn{6}{c}{\textbf{Closed-Source Large Language Models}}\\
\midrule

Gemini-2.5-flash   & 38.9 ± 2.4   & 55.0 ± 1.8   & 63.2 ± 0.7   & 85.0 ± 0.8   & 77.0 ± 0.8   \\
Gemini-3-flash     & 56.0 ± 1.6   & 62.9 ± 1.2   & 64.2 ± 1.2   & 77.4 ± 1.6   & 75.8 ± 1.8   \\
Gpt-4o-mini        & 34.9 ± 0.6   & 53.0 ± 0.8   & 60.6 ± 0.7   &  \textbf{99.3 ± 0.1}   & 92.0 ± 0.2   \\
GPT-4.1-mini       & 70.5 ± 1.0   & 79.2 ± 0.5   & 84.9 ± 0.3   & 99.0 ± 0.3   & 94.3 ± 0.3   \\
GPT-5-mini & 55.9 ± 1.0 & 61.8 ± 0.7 & 63.3 ± 0.5 & 88.8 ± 0.7 & 87.9 ± 0.9 \\
Qwen-flash         & 75.4 ± 0.4   & 81.3 ± 0.1   & 85.4 ± 0.3   & 98.6 ± 0.4   & 96.4 ± 0.5   \\
\midrule
\multicolumn{6}{c}{\textbf{Open-Source Large Language Models}}\\
\midrule
Qwen3-0.6B & 11.8 ± 1.3   & 33.6 ± 1.3   & 55.1 ± 1.0   & 95.5 ± 0.1   & 73.1 ± 0.8   \\
Qwen3-1.7B & 22.4 ± 2.3   & 47.6 ± 0.7   & 62.8 ± 0.5   & 96.4 ± 0.5   & 82.1 ± 1.0   \\
Qwen3-4B  & 36.7 ± 1.0   & 53.2 ± 1.4   & 62.1 ± 2.2   & 95.1 ± 0.7   & 86.6 ± 1.0   \\
Qwen3-8B & 42.4 ± 1.7   & 59.9 ± 2.1   & 69.8 ± 3.0   & 95.2 ± 1.1   & 86.6 ± 0.9   \\
DeepSeek-R1-Qwen3-8B & 32.7 ± 2.3   & 47.1 ± 2.7   & 56.0 ± 2.8   & 80.7 ± 2.6   & 71.6 ± 2.0   \\
DeepSeek-R1-Llama-8B & 27.9 ± 0.4   & 44.4 ± 1.2   & 54.7 ± 1.5   & 97.8 ± 0.5   & 88.0 ± 1.4   \\

\midrule
\multicolumn{6}{c}{\textbf{Ours}}\\
\midrule

SFT-Qwen3-4B  & 50.5 ± 0.2   & 68.8 ± 0.3   & 82.3 ± 0.3   &  95.7 ± 0.2 & 86.4 ± 0.1   \\

GRPO-Qwen3-4B & 51.9 ± 1.3 & 71.0 ± 0.9 & 86.7 ± 0.8 & 96.8 ± 0.1 & 84.5 ± 0.3 \\
SFT-GRPO-Qwen3-4B& 76.9 ± 1.5  & 85.4 ± 0.8   & 92.6 ± 0.1   & 96.0 ± 0.6   & 94.9 ± 0.4   \\
\model & \textbf{86.1 ± 0.7}   &  \textbf{90.6 ± 0.4}   &  \textbf{94.7 ± 0.4}   &  98.3 ± 0.3 &  \textbf{97.0 ± 0.3}   \\
\bottomrule
\end{tabular}
\caption{Experimental results on offline EQS-Benchmark. We compare our \model{} with strong baselines that satisfy the RT requirements. Each experiment was repeated 3 times; mean ± std reported.
\label{tab:main_result}}
\vspace{-0.9em}
\end{table*}


\subsection{Main Experiments}
We evaluate different baselines in both offline and online experiments. Results demonstrate the effectiveness of our \model, and further reveal a strong positive correlation between offline and online performance. 

\subsubsection{Offline Experiments}
As shown in the Table \ref{tab:main_result}, our proposed framework significantly outperforms existing baselines. Our final model \model{} achieves 86.1\% ± 0.7\% Strict Accuracy and 90.6\% ± 0.4\% \sacc, substantially surpassing both the backbone Qwen3-4B ( \qacc{} 36.7\%, \sacc{} 53.2\%) and the strong closed-source baseline qwen-flash (\qacc{} 75.4\%, \sacc{} 81.3\%), with +10.7\% and +9.3\% gains in \qacc{} and \sacc{}, respectively. 

\paragraph{Effect of group-disagreement-driven mining.}
Compared to SFT-GRPO-Qwen3-4B (\qacc{} 76.9\%, \sacc{} 85.4\%), \model{} improves \qacc{} by 9.2\% and \sacc{} by 5.2\%. This suggests that group-disagreement-driven mining effectively surfaces hard query contexts. By prioritizing such informative contexts during RL, \model{} achieves better generalization and stronger robustness.

\paragraph{Effect of actionable quality reward.} Even without warm-start stage, applying only our actionable quality reward in RL fine-tuning (GRPO-Qwen3-4B) boosts Qwen3-4B's \qacc{} from 36.7\% to 51.9\% (+15.2\%) and \sacc{} from 53.2\% to 71.0\% (+17.8\%). This demonstrates the effectiveness of our actionable quality reward even without the SFT warm-up stage.

\begin{figure}

  \centering
\subfigure[Online.]{
   \label{fig:5_1}
   \vspace{-0.9em}
  \includegraphics[scale = 0.38]{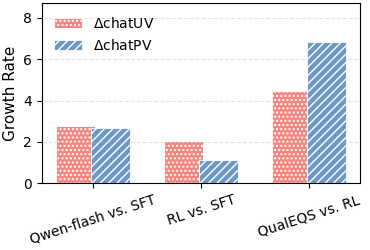}}
  \vspace{-0.9em}
\subfigure[Offline.]{
  \label{fig:5_2}
  \includegraphics[scale = 0.38]{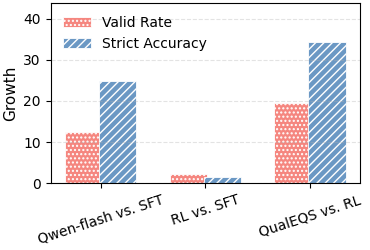}}
\caption{Comparison of online and offline experiments.} 
  \label{fig:online} 
\vspace{-0.9em}
\end{figure}

\subsubsection{Online Experiments}
Figure~\ref{fig:online} reports the $\Delta$ChatUV and $\Delta$ChatPV results across the online experimental groups. Compared with the GRPO-based baseline, our \model{} improves $\Delta$ChatPV by 6.81\%, demonstrating its effectiveness in guiding users toward more engaging conversations. The results across the three experiment groups also show a consistent trend between offline and online evaluations: models with higher Strict Accuracy achieve better online performance.

Beyond these engagement metrics, we further analyze several transaction-related online metrics that are particularly important in the conversational shopping setting. Since product recommendation is a core user intent in our scenario, the quality of downstream interactions can also be reflected by the number of buyers, the number of orders, and Gross Merchandise Volume (GMV). Compared with the RL baseline, our \model{} delivers substantial gains on three metrics, with improvements of 6.24\% in the number of buyers, 8.47\% in the number of orders, and 12.81\% in GMV, further demonstrating its practical business value.

\subsection{Further Analysis}
\paragraph{Ablation Study.} We perform an ablation study to further analyze the effectiveness of each component in our framework. As shown in Table~\ref{tab:abltion}, when we remove group-disagreement-driven mining and instead use all non-click data in the RL stage, the \qacc{} drops by 4.5\%. Additionally, removing the mixed-data ratio used in the warm-start stage leads to a 3.9\% \qacc{} drop. 

\begin{figure}[h]
	\centering
	\includegraphics[width=0.22\textheight]{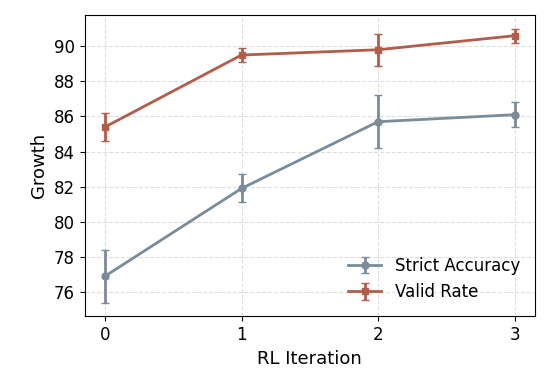}
	\caption{Effect of Iteration.}
	\label{fig:iteration}
    \vspace{-0.9em}
\end{figure}

\paragraph{Effect of Iterative Refinement.} 
We also analyze the effect of iteration. As shown in Figure \ref{fig:iteration}, performance improves as the number of iterations increases, converging at $k = 3$, which demonstrates the effectiveness of our iterative strategy.

\begin{table}
\centering \small
\begin{tabular}{l c c  }
\toprule

 Models&\qacc & \sacc \\
 \midrule
\model & 86.1 ± 0.7   & 90.6 ± 0.4 \\
 w/o disagreement & 81.6 ± 0.9   & 88.4 ± 0.5\\

 \midrule
SFT-Qwen3-4B & 50.5 ± 0.2   & 68.8 ± 0.3\\
w/o mix ratio & 46.6 ± 2.0 & 66.1 ± 1.1 \\
\bottomrule
\end{tabular}
\caption{\label{tab:abltion}Ablation Study.
}
\vspace{-0.9em}
\end{table}

\begin{figure}
	\centering
	\includegraphics[width=0.22\textheight]{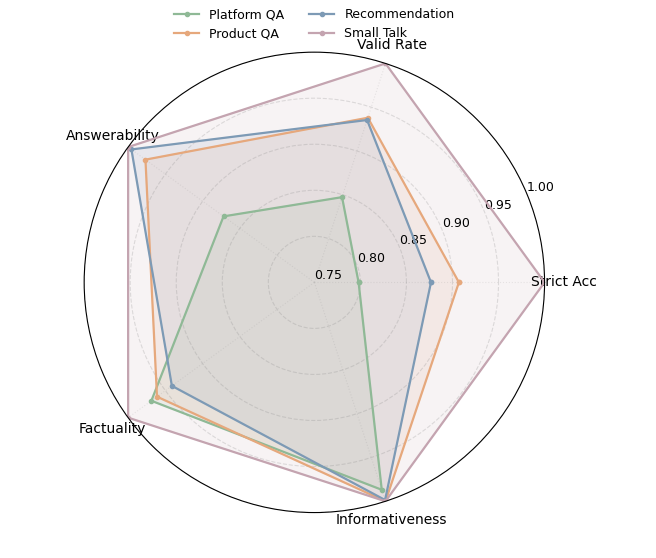}
	\caption{Experimental results of \model{} across four e-commerce intents.}
	\label{fig:intent}  
 \vspace{-0.9em}
\end{figure}

\paragraph{Effect Across Different Intents.} We further evaluate \model{} across four intent types. As shown in Figure~\ref{fig:intent}, \model{} performs strongly on product recommendation, product QA, and small talk, but remains weaker on platform QA. We attribute this gap to the higher demand for platform-specific knowledge and stricter answerability constraints, suggesting a direction for future improvement.

\section{Related Work}
Existing query suggestion methods mainly can be divided into two categories: retrieval-based \cite{retrieval2018entity,retrieval2008context,retrieval2023exploiting} and generation-based approaches \cite{Min2025ctr,bi2025consistency,Sun2026QueryAgentR1BQ,zheng2026icebreakerconversationalagentsbreaking}. Retrieval-based methods recall candidate queries from a historical query pool that are similar to the current user query, typically leveraging user behavior signals such as co-occurrence. However, these methods are inherently limited by their reliance on pre-existing query patterns and struggle to capture nuanced, real-time user intent \cite{smith2018understanding,gal2017deep,van2020uncertainty,gal2016dropout}.

Recently, research efforts have shifted toward generation-based approaches, which can dynamically recognize the user's latest intent and contextual information to directly generate diverse and relevant query suggestions \cite{bacciu2024generating,shen2024enhancing,wang2023zero}. 
For instance, \citet{Min2025ctr} proposed a method that leverages a click-through rate (CTR) model as a reward signal to construct preference pairs from generated queries, then applies Direct Preference Optimization (DPO) \cite{dpo} to fine-tune an LLM for query suggestion. \citet{bi2025consistency} incorporated both the CTR model and a topic expansion reward during preference pair construction, aiming to better align suggested queries with user interests and topic expansion. \citet{yiyi2025clicks} proposed a multi-stage framework designed to achieve progressive alignment between the generation policy and user preferences. However, the above methods heavily rely on abundant, high-quality online user click data to train reliable CTR models \cite{guo2021embedding,Wang_Yang_Chen_Liu_Zhao_Guo_Sun_Li_2021}, which is particularly challenging in early-stage deployment scenarios \cite{kim2024improving,schein2002methods,lika2014facing}. 

\section{Conclusion}
In this work, we present \model{}, a quality-first iterative RL framework for e-commerce query suggestion under sparse feedback. \model{} optimizes actionable suggestion quality with answerability, factuality, and information gain, and leverages group-level reward disagreement to mine hard online contexts for iterative improvement. We also introduce EQS-Benchmark and show that quality-based offline metrics strongly correlate with online performance. Results in both offline and online settings demonstrate that \model{} consistently outperforms strong baselines, with a 6.81\% gain in online ChatPV.

\section*{Ethical Considerations}
In deploying our query suggestion model, we adhere to two core ethical principles. (1) Data Privacy Protection: All training and inference data are strictly de-identified and do not contain any personally identifiable information, user platform IDs, or other private attributes. The model operates solely on anonymized interaction logs, ensuring user privacy is preserved throughout the pipeline. (2) Content Safety Risks: Sensitive topics are explicitly excluded during both training and inference stages to prevent the generation of inappropriate or harmful suggestions. This proactive filtering strategy significantly mitigates potential risks associated with biased or unsafe content.

\section*{Limitations}
Although \model{} brings consistent improvements across different intent types under sparse-feedback settings, platform QA still shows more room for improvement than product recommendation, product QA, and small talk. Compared with the other intents, platform QA often involves questions that depend more heavily on platform rules, service details, and policy-related knowledge, while also requiring stricter answerability control. In the future, we plan to address this by incorporating richer platform question-answering knowledge and stronger answerability constraints.

\bibliography{custom}

\appendix
\definecolor{codebg}{rgb}{0.96,0.96,0.96}
\lstdefinestyle{promptstyle}{
    basicstyle=\ttfamily\small,
    breaklines=true,
    frame=single,
    backgroundcolor=\color{codebg},
    rulecolor=\color{gray!30},
    xleftmargin=1em,
    xrightmargin=1em,
    showstringspaces=false,
    tabsize=4,
    keepspaces=true
}

\section*{Appendix} 
\label{sec:appendix}

\section{Prompt for General Intent}
\label{sec:appendix_A}

\begin{lstlisting}[style=promptstyle]
# Task: identify the user's intention based on the conversation with the AI assistant and Generate 3 follow-up options for the user to click on:
- If the input contains errors or intention is unclear, infer 3 probable questions.
- If intention is clear but broad, narrow the scope by asking about specific details or next steps.
- If intention is clear and specific, broaden the scope by asking about complementary, alternative, or exploratory questions.

# Constraints
1. Each follow-up option must be concise (under 30 tokens) and unique.
2. When generating options, you should from the user's perspective.
3. Present the output in the following JSON structure:
{
"options": [
    "Follow-up option 1",
    "Follow-up option 2",
    "Follow-up option 3"
    ]
}

\end{lstlisting}

\section{Prompt for Product Recommendation}
\label{sec:appendix_B}

\begin{lstlisting}[style=promptstyle]
# Task
Identify the user's intention based on the conversation with the AI Shopping assistant:

Unclear
Clear but broad
Clear and specific
Generate 3 follow-up options for the user to click on:

If the input contains errors or the intention is unclear, infer and generate probable user intentions, then phrase three as user-initiated questions or requests to the shopping assistant.
If the intention is clear but broad, help narrow the scope by generating three user-initiated questions that ask for specific details or next steps.
If the intention is clear and specific, help broaden the scope by generating three user-initiated questions about complementary, alternative, or exploratory topics.
# Constraints
Each follow-up option must be concise (under 30 tokens) and unique.
All follow-up options must be written in first person, as natural questions or statements that the user would click on or say to the shopping assistant.
Do not use sentences from the assistant's perspective. Only output the user's words.
For example:
"Can you recommend something popular?"
"Are there any deals on electronics today?"
"I'm looking for a birthday gift. Any suggestions?"
If the user's input is a broad product category (e.g., "flowers"), use contextual clues to infer relevant subcategories or usage scenarios, and phrase follow-ups accordingly (e.g., "Which flowers are best for gifting elderly people?")
Do not include multiple choices in a single option. Each option must be a single question or request.
If multiple product-related queries exist in the conversation history, only generate follow-ups for the most recent product query.
Avoid questions that compare non-similar products.
If a follow-up question involves budget, use a specific numerical amount instead of vague terms.

Important:
Always output follow-up options as if the user is talking or clicking, never as assistant suggestions.
If the intention is unclear, do not repeat or paraphrase the user's vague input; instead, infer possible intended scenarios and phrase each as a user-initiated question or request.
If a follow-up question involves budget, be mindful of reasonable price ranges for products. Use the currency specified by the user or their country, and prioritize any input restrictions or special requirements when displaying prices.

Present the output in the following JSON format:

Example input:
user query: Buy something
{
"options": [
    "Can you show me popular items right now?",
    "Can you recommend good daily necessities?",
    "I'm buying a gift for a friend. Any suggestions?"]
}

\end{lstlisting}

\section{Prompt for LLM-as-Judge}
\label{sec:appendix_C}

\begin{lstlisting}[style=promptstyle]
# Task: Evaluate follow-up question quality for (Anonymous) e-commerce assistant
You are a professional e-commerce follow-up question quality evaluator, tasked with strictly assessing whether the three follow-up questions provided by the user meet (Anonymous)'s high-quality standards.

Evaluate each of the three follow-up questions against the following three rules. If any single follow-up question violates any one rule, it is considered a bad case.  

1. Answerability 
- The question must be a genuine product-related or platform QA inquiry or Common Sense Chat that a real user would ask within the (Anonymous) app.  
- Prohibited content includes AI/system-perspective counter-questions (e.g., "What would you like to know?"), privacy-related inquiries (e.g., "What kind of flowers does your girlfriend like?"), or questions outside the scope of common e-commerce knowledge (e.g., weather forecasts, medical advice).  

Compliant examples:  
- "Does this Bluetooth headset support active noise cancellation?"  
- "Is the inner pot of this rice cooker made of stainless steel or coated material?"  

Non-compliant examples:  
- "What type of product do you want to buy?"  
- "What color of flowers does your girlfriend like?"  
- "Can you tell me if it will rain tomorrow?"  

2.Factual Accuracy
- The content of the question must align with real-world fact and must not contain clearly fictional or impossible statements.  

Compliant examples:  
- "Which Type-C cables support fast charging?"  
- "Can you recommend an Android phone with a battery capacity above 5000mAh?"  

Non-compliant examples:  
- "Is there an iPhone 15 for 1 PHP?"  
- "Can I buy a flying electric scooter?"  
- "Are there wireless earbuds that don't require charging?" 
- "Does (Anonymous) sell time machines?"  

3. Information Gain
- Each follow-up question must deepen, refine, or clarify the original intent (e.g., by specifying brand, features, budget, etc.).  
- If a budget is mentioned, it must include a specific numeric value (e.g., "under 2,500 PHP"); vague terms such as "cheap" or "reasonable" are prohibited.  
- The question should be related to user intent but must not merely rephrase or repeat the original query without adding new information.  

Compliant examples:  
- Original: "I want to buy a router."  Follow-up: "Do you have any Xiaomi routers under 2,500 PHP that support Wi-Fi 6?"  
- Original: "Recommend an electric toothbrush."  Follow-up: "Are there any electric toothbrushes with pressure sensors under 2,000 PHP?"  
- Original: "How is this hair dryer?"  Follow-ups: "What is its wattage?", "Does it support cold/hot air switching?"  
Non-compliant examples:  
- "Can you recommend a router?" (no added information)  
- "Are there cheaper phones?" (vague budget)  
- "Is this headset good?" (lacks specific criteria)  
- "Recommend another electric toothbrush." (mere repetition of request)  

Output Requirements
- Output only a valid JSON object.  
- If all three follow-up questions fully comply, return {"score": 3,"reason":"reason for score", "Answerability":3,"Factual_Accuracy":3,"Information_Gain":3}.  
- If exactly one question violates any rule, overall score got 2; if two violate,  overall score got 1; if all three violate,  overall score got 0.  
- Do not include explanations, comments, or any additional text.
\end{lstlisting}

\section{Test Set Construction}
\label{sec:benchmark}
To construct a high-quality benchmark test set, we first collected approximately 1,500 raw user queries from four core intent categories: small talk, product recommendation, product QA, and general platform QA. From each category, we initially sampled 100 candidate utterances, ensuring coverage across diverse usage scenarios.

This initial pool underwent a rigorous manual curation process to eliminate noisy or low-quality samples that could introduce bias or ambiguity into the evaluation. Annotators were instructed to exclude instances that exhibited any of the following characteristics: semantic redundancy or high similarity to other queries (to avoid evaluation bias from near-duplicate intents); unclear or unintelligible semantics; severe spelling or grammatical errors that impede comprehension; garbled characters, meaningless symbols, or standalone emoji sequences without textual context; and fragmentary inputs lacking a clear user intent, such as single characters, isolated words, or phrases devoid of contextual grounding.

The curation was performed by trained annotators through multiple rounds of review, with disagreements resolved via consensus. After iterative filtering and quality control, the final benchmark test set comprises 500 high-quality, semantically coherent, and intent-unambiguous samples.

\section{Reward Model Deployment}
\label{sec:reward_deployment}
We continuously monitor the reward model service throughout training. The service sustains an average query throughput of 23.8 QPS with an average response latency of 1.15 seconds. Because the training pipeline is asynchronous, reward queries are processed in parallel with rollout generation, the reward model service does not constitute a practical bottleneck. We also track service failures during RL training. The average failure throughput is 0.434 QPS, corresponding to an overall failure rate of approximately 1.8\%. To improve robustness, we employ two safeguards. First, each failed reward query is automatically retried up to three times to mitigate transient issues such as network jitter or brief service interruptions. Second, if all retries fail, we assign a neutral reward of 0 to the corresponding sample, thereby reducing the impact of unreliable reward signals and helping maintain training stability.

\section{Online Metric}
\label{sec:online_mtric}

In our online A/B experiments, we evaluate the impact of different QS models on user engagement with the in-app chat functionality using two primary metrics:

\begin{itemize}
    \item \textbf{ChatUV (Chat Unique Visitors)}: The number of distinct users who \textit{initiated or participated in at least one chat conversation} during the observation period. Each user is counted only once, regardless of how many conversations they engaged in.
    
    \item \textbf{ChatPV (Chat Participation Views)}: The total number of chat-related interactions (e.g., sending messages, replying, etc.) generated by all users during the same period. This metric is not deduplicated; thus, repeated participations by the same user are counted multiple times.
\end{itemize}

To quantify the relative improvement of the treatment group (denoted as $A$) over the control group (denoted as $B$), we compute the \textbf{Growth Gap} as follows:
\begin{equation}
    \text{Growth Gap} = \frac{A - B}{B},
\end{equation}
where $A$ and $B$ represent the observed values of either ChatUV or ChatPV in the treatment and control groups, respectively. The Growth Gap expresses the percentage change attributable to the experimental intervention. A positive value indicates an increase in user engagement due to the treatment, while a negative value suggests a decline.

\section{Training Detail}
\label{sec:appendix_D}
All experiments were conducted on 8 NVIDIA A100 80GB GPUs. We first performed SFT on Qwen3-4B\cite{yang2025qwen3technicalreport} using the LLaMA-Factory \cite{zheng2024llamafactoryunifiedefficientfinetuning} framework with full-parameter updates, a maximum sequence length of 4096, an effective batch size of 256, a learning rate of 1e-4, bf16 precision, and 3 epochs of training. To mitigate catastrophic forgetting, the warm-start stage employed a mixed-data strategy, training on a proportionally combined dataset that concatenates the target-domain data with the general-purpose QA dataset ShareGPT4 \cite{chen2023sharegpt4vimprovinglargemultimodal}. 

The resulting model was then used as the initial policy for reinforcement learning via GRPO in the VeRL \cite{Sheng_2025} framework. In the GRPO stage, each rollout response was scored by a custom reward function that leverages an LLM-as-judge \cite{li2025generationjudgmentopportunitieschallenges} to obtain reward score. Training was configured with a batch size of 256, prompt and response length capped at 2048 and 512 tokens respectively, vLLM-based rollouts, an actor learning rate of 1e-6, and a KL loss coefficient of 0.001.

\end{document}